\documentclass[letterpaper, 10 pt, conference]{ieeeconf}  

\IEEEoverridecommandlockouts                              

\usepackage{graphics} 
\usepackage{epsfig} 
\usepackage{mathptmx} 
\usepackage{times} 
\usepackage{amsmath} 
\usepackage{amssymb}  
\usepackage{subfigure}
\usepackage{makecell}
\usepackage{verbatim} 
\usepackage{hyperref} 
\usepackage{cite}
\usepackage{stfloats}
\usepackage{algpseudocode}
\usepackage[ruled,vlined]{algorithm2e}
\usepackage[table]{xcolor}
\usepackage{multirow}
\usepackage{booktabs} 
\definecolor{ForestGreen}{rgb}{0,0.6,0}

\DeclareMathAlphabet{\mathcal}{OMS}{cmsy}{m}{n}
\SetMathAlphabet{\mathcal}{bold}{OMS}{cmsy}{b}{n}

\title{\LARGE \bf
CNN-based Omnidirectional Object Detection for HermesBot Autonomous Delivery Robot with Preliminary Frame Classification
}

\author{Saian Protasov, Pavel Karpyshev, Ivan Kalinov, Pavel Kopanev,    \\ Nikita Mikhailovskiy, Alexander Sedunin, and Dzmitry Tsetserukou
\thanks{All authors are with Space Center of Skolkovo Institute of Science and Technology, Moscow, Russia 
        {\tt ivan.kalinov@skolkovotech.ru, \{saian.protasov, pavel.karpyshev, pavel.kopanev, nikita.mikhailovskiy, alexander.sedunin, d.tsetserukou\} @skoltech.ru}}
}

\begin{document}

\maketitle
\thispagestyle{empty}
\pagestyle{empty}

\begin{abstract}

Mobile autonomous robots include numerous sensors for environment perception. Cameras are an essential tool for robot's localization, navigation, and obstacle avoidance. To process a large flow of data from the sensors, it is necessary to optimize algorithms, or to utilize substantial computational power. In our work, we propose an algorithm for optimizing a neural network for object detection using preliminary binary frame classification. An autonomous outdoor mobile robot with 6 rolling-shutter cameras on the perimeter providing a 360-degree field of view was used as the experimental setup. The obtained experimental results revealed that the proposed optimization accelerates the inference time of the neural network in the cases with up to 5 out of 6 cameras containing target objects.

\end{abstract}

\section{Introduction}
\subsection{Motivation}
During recent years, the e-commerce market has been growing at an astounding speed. According to \cite{tmirob}, it is expected to grow up to more than EUR 2.5 trillion by the year 2023. Such rapid growth has inevitably led to the gain in the global delivery market. One of the key parts of delivery services is last-mile delivery from the nearest warehouse or retail store directly to the customer’s door. This part of the delivery process directly influences customer satisfaction, which is essential for logistic companies' success. Such delivery was typically performed using vehicles. However, due to traffic congestion the delivery time using vehicles has significantly increased \cite{akeb2018building}, thus forcing logistics providers to develop new methods of last-mile delivery that rely less on traffic situation. For example, the combination of delivery vans and electric bikes has shown significant increase in delivery efficiency \cite{melo2017evaluating}.

The Covid-19 pandemic of 2020 has created a rapidly growing trend for minimizing personal contact and robotic automatization in all areas of human life and industry application, e.g., for stocktaking automation in warehouses \cite{kalinov2019high, kalinov2020warevision,kalinov2021impedance, kalinov2021warevr}, shopping malls \cite{petrovsky2020customer}, disinfection of hospitals and offices \cite{perminov2021ultrabot,mikhailovskiy2021ultrabot}, and autonomous chargers \cite{okunevich2021deltacharger}. For epidemiological security, and due to vast advances in automation technology, various contact-less last-mile delivery methods are currently being developed. According to \cite{gromls}, up to 80\% of Business to Customer (B2C) deliveries can be automated. 

Multiple large companies, such as Amazon, Google, DHL, UPS etc. are already testing automated delivery using Unmanned Aerial Vehicles (UAVs). However, the use of UAVs is limited by their low lifting capacity, range, and cost-effectiveness. The aerial vehicles can also become a threat in case of malfunction, and are prohibited in urban areas of certain countries. For these reasons, the use of Autonomous Ground Vehicles (AGV) has been extensively researched as well. Several companies, such as Starship Technologies, ANYmal and FedEX, are already developing ground robots for last-mile delivery.
The characteristics of aforementioned robots make them perfect for short-distance deliveries in urban areas. Namely, the Starship Technologies robot is a 6-wheeled platform that weighs around 45 kilograms and is capable of moving within a 5-kilometer radius at pedestrian speed with 2.6 kg of payload \cite{kotasova}. Other approaches to autonomous last-mile delivery were introduced by the ANYmal team with their four-legged robot and the FedEX Bot.


Despite multiple robots being developed at the moment, the autonomous last-mile delivery AGVs still face numerous difficulties. Such machines need robust and precise localization technologies, as well as reliable obstacle detection and avoidance algorithms.

\begin{figure} [!t] 
\begin{center}
\includegraphics[width=8 cm]{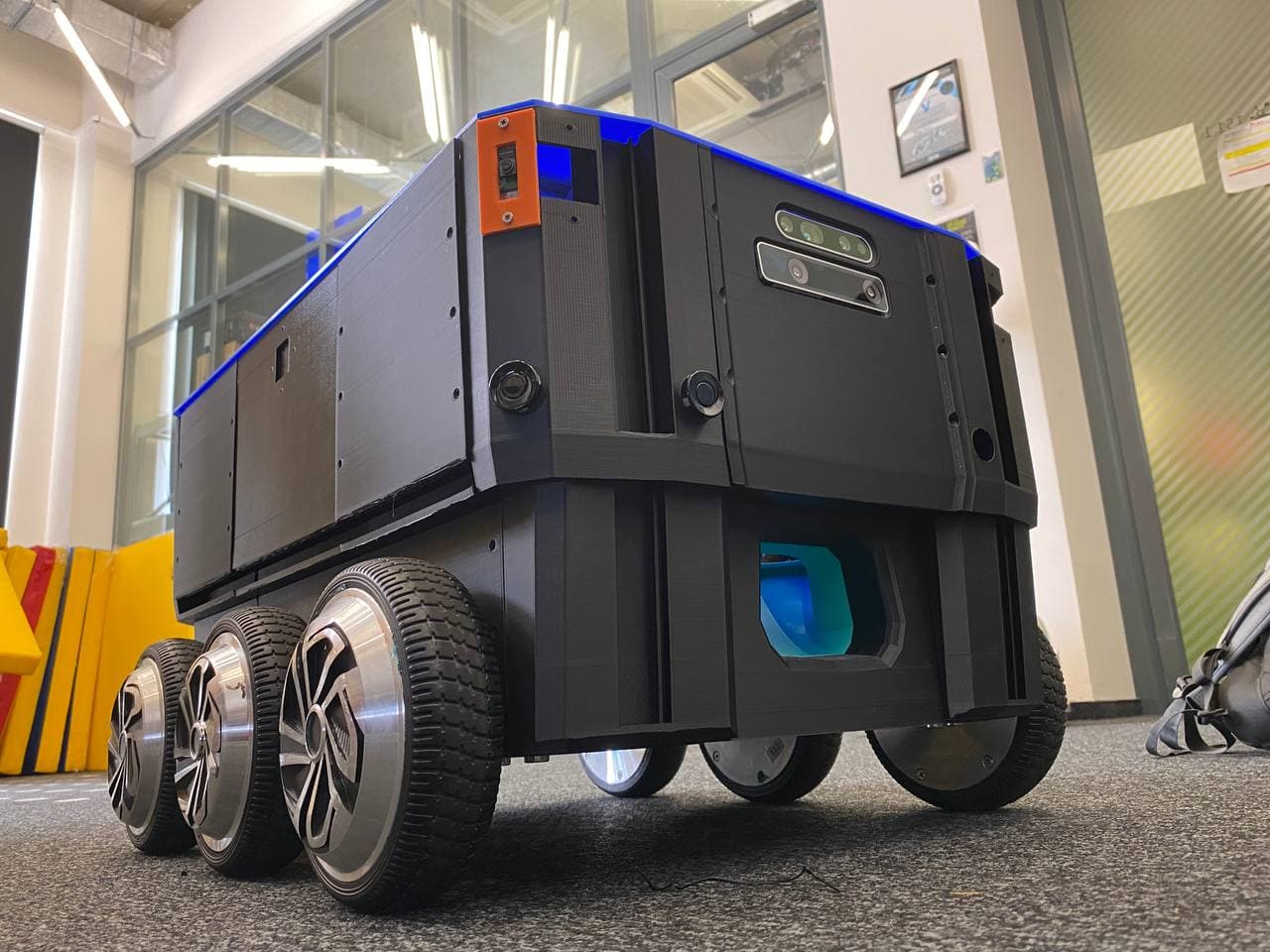}
\caption{HermesBot outdoor delivery platform}
\vspace{-2.5em}
\label{fig1}
\end{center}
\end{figure}

\subsection{Problem Statement}
For obstacle detection and avoidance, the robot needs precise and real-time information about its surroundings. This can be achieved using visible- and infrared-spectrum cameras, LiDARs, ultrasonic distance sensors, and other types of data acquisition devices. The data flow from all these devices is huge even for desktop computers, not to mention small and low-power mobile devices. Increasing the computational power of a mobile robot results in increased energy consumption, and thus, to limited operation time, working range, and cost efficiency. Because of this, mobile robots are often subject to trade-off between accuracy and working time. The particular example of such task is pedestrian detection. For this purpose, the robots are usually equipped with multiple visible range cameras faced in all directions, thus allowing for 360-degree field of view to detect surrounding pedestrians and perform necessary avoidance maneuvers. For example, the robot by Starship technologies is equipped with 9 cameras, with 3 facing in the front direction, 2 in the rear direction and 4 on the sides. These cameras create a data flow which is extremely difficult for mobile computing units to process.

\subsection{Related works}
Object detection algorithms are widely used in the area of mobile robotics. Typically, objects to detect depend on the task that the robot is performing. For example, P. Karpyshev et al. \cite{karpyshev2021autonomous}  successfully used the Mask-RCNN neural network for apple trees disease detection. I. Kalinov et al. \cite{kalinov2020warevision} presented a heterogeneous robotic system with a different task — the detection of barcodes during warehouse stocktaking using UAVs. In this case, the authors used the U-Net convolutional neural network for the detection and semantic segmentation of barcodes. Patrick K. Chemeli et al. \cite{roboarm} used the Single Shot Detector to locate and identify objects using a robotic arm. All these tasks cannot be solved from the point of view of optimizing neural network algorithms. In the cases described above, the robots either had few cameras, or there were enough computing resources for prompt detection, such as dedicated desktop-grade GPUs. Bernd Poppinga et al. \cite{jetnet} offered an ultra-lightweight architecture for object detection during robotic soccer competitions. However, such architectures are often used for very specific tasks under certain conditions. In this article, we propose a method to increase the speed of an object detection neural network for a robot with a high number of cameras.

The efficiency of using neural networks in the detection task is an indisputable fact. At the moment, there is a large number of different neural network architectures for object detection. All detectors can be divided into 2 types: two-stage detectors and one-stage detectors.

One of the most popular two-stage architectures for object detection is Faster RCNN \cite{faster}, which is a continuation of the ideas in Fast RCNN \cite{fast} and RCNN \cite{rcnn}. The Faster R-CNN architecture is formed as follows: An image is fed to the input of a convolutional neural network, where a feature map is formed. The feature map is processed by the RPN layer: a sliding window is traversed over the feature map. The center of the sliding window is linked to the center of the anchors, areas that have different aspect ratios and different sizes. The authors use 3 aspect ratios and 3 sizes. Based on the intersection-over-union (IoU) metric, the degree of intersection of anchors and true marked rectangles, a decision is made about the current region — whether there is an object in it or not. Next, the FastCNN algorithm is used: the feature map with the obtained objects is transferred to the RoI layer, followed by the processing by fully connected layers and classification, as well as determining the displacement of the potential objects' regions.

One-stage detectors work in a different way. The class and coordinates are predicted in one step from the anchor bounding boxes that tightly cover the picture with different scales and different aspect ratios. The representatives of this class are SSD \cite{ssd} and YOLO \cite{yolov3}. A one-stage detector only requires one pass through the neural network and predicts all bounding boxes in one go. It is much faster than two-stage detectors and is more suitable for mobile devices.

\begin{figure} [!t]
\centering
\includegraphics[width=0.9\linewidth]{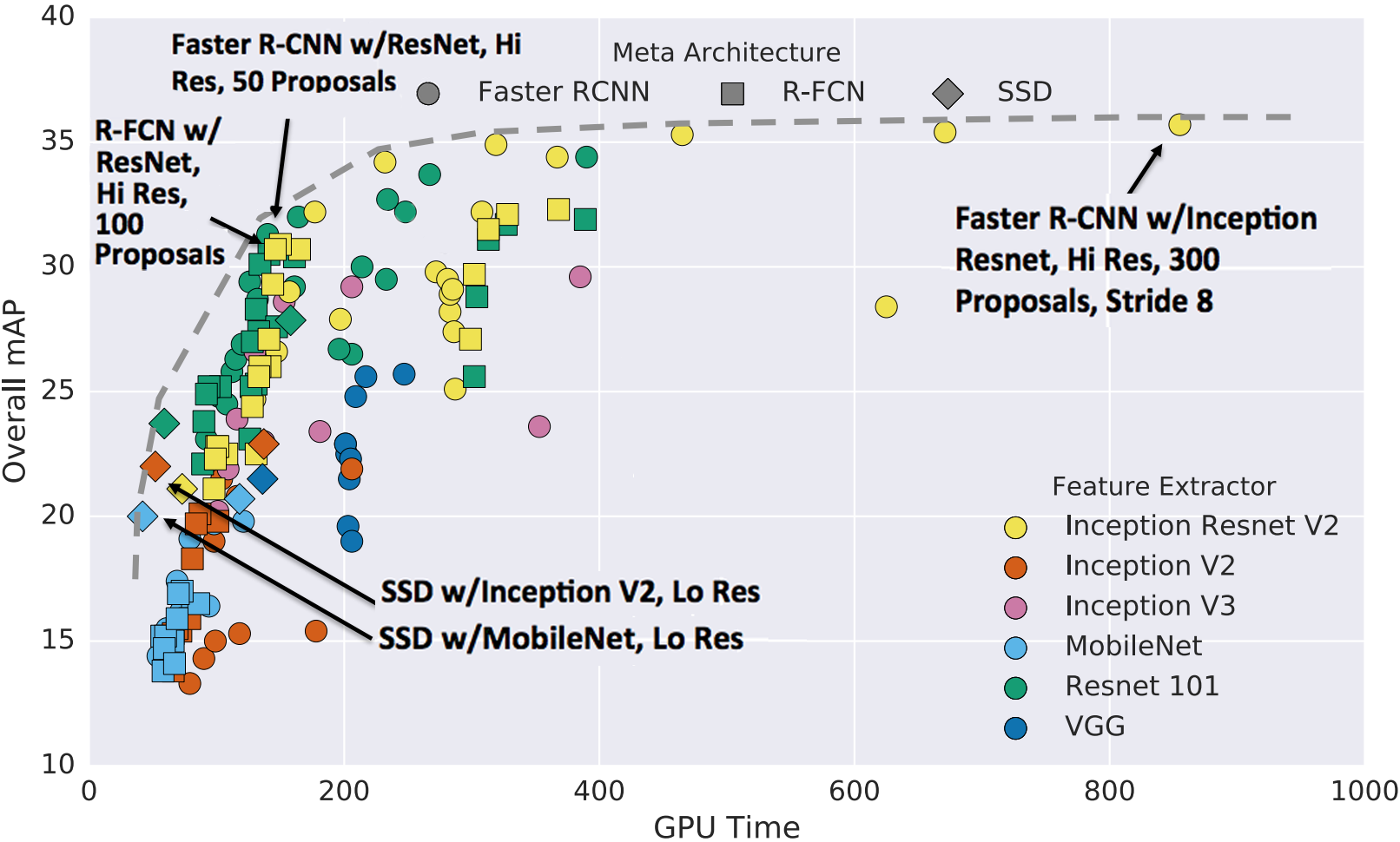}
\vspace{-1em}
\caption{Accuracy vs time, with marker shapes indicating meta-architecture and colors indicating feature extractors. Each (meta-architecture, feature extractor) pair can correspond to multiple points on this plot due to changing input sizes, stride, etc.} 
\vspace{-2em}
\label{fig:ggl}
\end{figure}

In the article by Jonathan Huang et al. \cite{google} multiple one-stage and two-stage detectors are compared using different metrics of accuracy and speed of execution (Fig. \ref{fig:ggl}). The authors conclude that SSD with MobileNet provides the best accuracy among the fastest detectors, and R-FCN \cite{rfcn} and SSD neural networks are faster on average, but cannot outperform Faster R-CNN in accuracy if speed is not the main goal.

\subsection{Contribution}
The proposed paper aims to improve the efficiency of object detection on systems with multiple cameras and limited computational power, i.e., mobile delivery robots. In the scope of this article, we use the HermesBot delivery robot, which includes six rolling shutter cameras for obstacle and pedestrian detection, as the hardware platform for experiments. The authors propose to use an additional lightweight classification network before the detection network in order to skip frames with no target objects present. In this case, in most situations, fewer calculations will have to be performed in order to get reliable information about target objects around the robot. The performed experiments have shown that if even one out of six cameras does not contain objects of interest, the proposed algorithm shows a significant increase in processing speed.

\section{System Overview}
\begin{figure} [!b] 
\vspace{-2em}
\begin{center}
\includegraphics[width=7.4 cm]{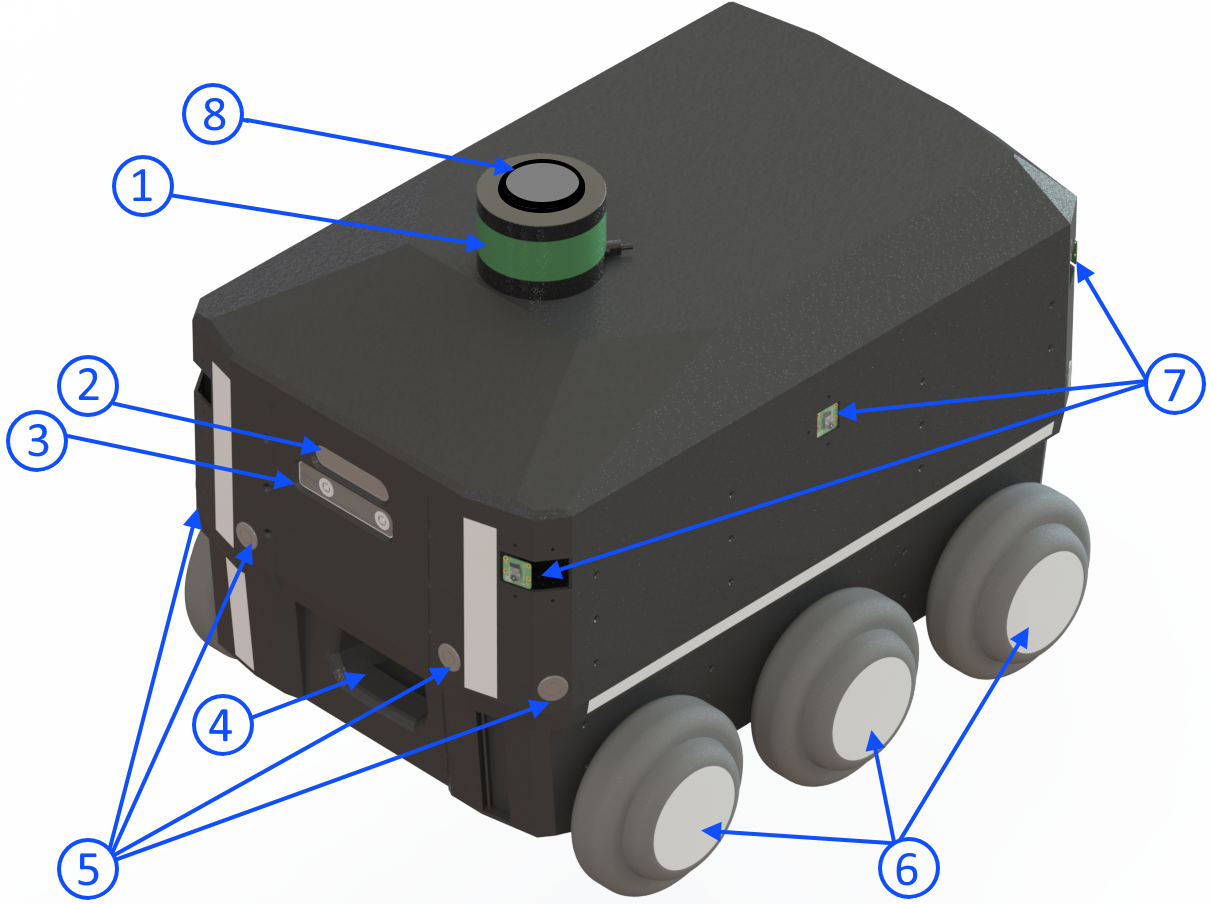}
\vspace{-1em}
\caption{Hardware equipment of HermesBot: 1. Velodyne VLP-16 3D LIDAR, 2. RealSense D435 depth camera, 3. RealSense T265 tracking cameras with IMU, 4. Vent, 5. Ultrasonic  Rangefinders SR04Tv3-US, 6. Wheel encoders with 3840 counts per revolution, 7. Rolling-shutter RasPi NoIR V2 camera with Sony IMX219 8-megapixel sensor. 8. Garmin GNSS receiver}
\label{fig2}
\end{center}
\end{figure}

Our team has developed a platform for autonomous delivery robot software research. The render of the robot is presented in Fig. \ref{fig2}. The platform is equipped with all hardware components necessary for outdoor movement, as well as software modules for movement control, path planning, SLAM, and obstacle detection and avoidance. The in-depth description of the robot is presented in sections below.

\subsection{Hardware}
The robot localization is mainly performed using two sets of RealSense cameras on the front and back sides of the platform: A RealSense D435 depth camera that is capable of acquiring 1280x720 depth data at up to 90 fps, and a RealSense T265 tracking camera with dual fisheye lenses and a high-precision IMU for acquiring high quality visual odometry. Pedestrian detection is performed using six RasPi NoIR V2 cameras located on both sides of the robot, facing sideways and at the angle of 45 degrees in all directions. Each camera has the maximum resolution of 3280 × 2464  and 62.2 degrees field of view. This allows for a 360-degree FoV and guarantees robust detection of stationary and moving obstacles. Close-range obstacle detection is performed using eight ultrasonic rangefinders operating at 10 Hz. Additional localization data is acquired from the Garmin GNSS receiver. Wheel odometry (WO) is collected using encoders with 3840 count per revolution. A VLP-16 3D LIDAR is installed on top of the platform for ground truth dataset collection. The robot also includes light sources.

All the calculations and robot control are performed using an Intel NUC8i7BEH computing module with Intel Core i7-8559U processor and 32Gb of RAM. Image processing and pedestrian detection is performed on an NVIDIA Jetson AGX Xavier computing unit with a 512-core Volta GPU containing Tensor Cores.

\subsection{Software}
\begin{figure} [h] 
\vspace{-0.5em}
\begin{center}
\includegraphics[width=7.4 cm]{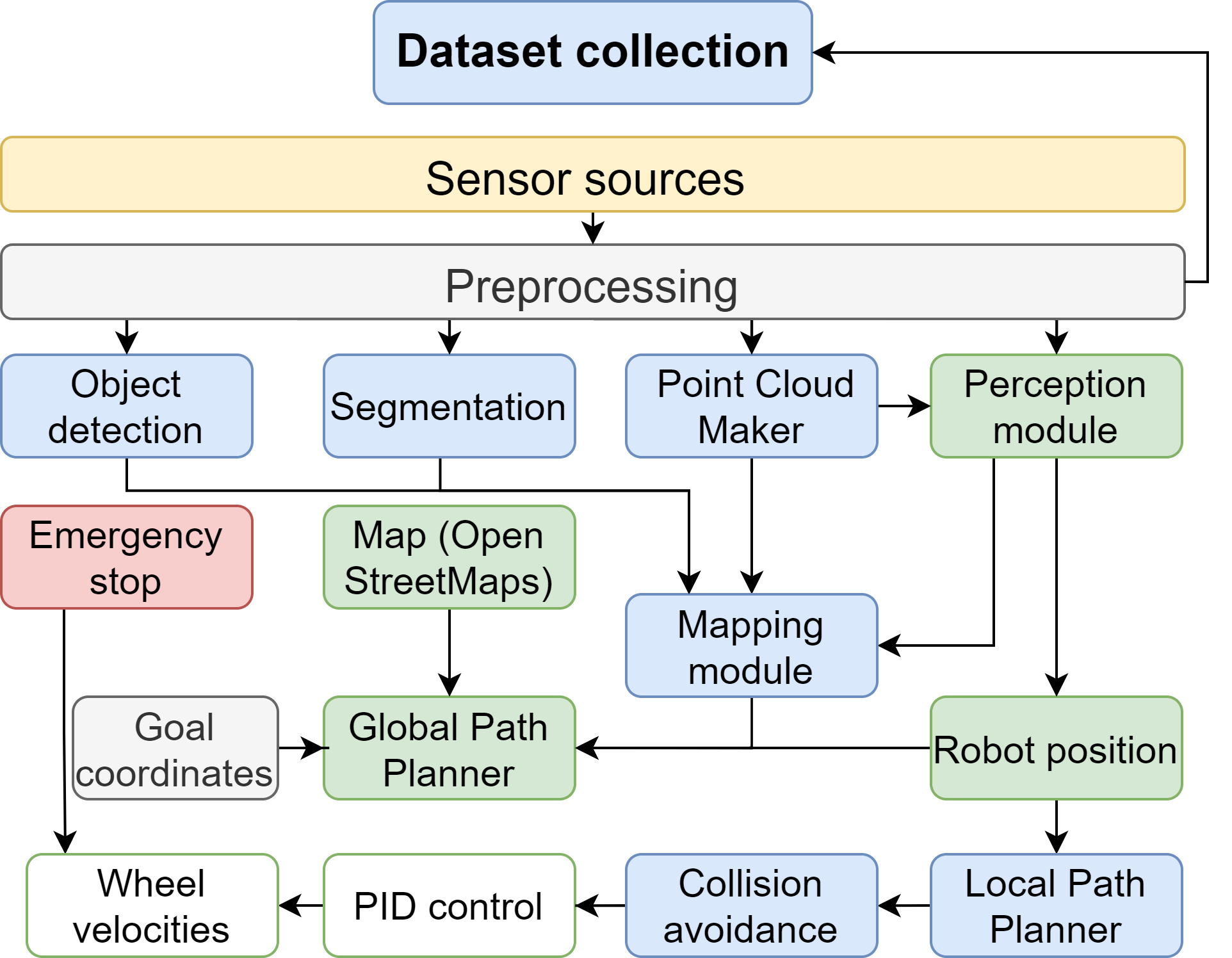}
\vspace{-0.5em}
\caption{Software architecture of HermesBot}
\vspace{-1.5em}
\label{fig3}
\end{center}
\end{figure}

The whole system architecture was developed with the use of the Robot Operating System (ROS). The schematic representation of the software architecture is presented in Fig. \ref{fig3}. The data from all sensors is sent to the Intel NUC computing module, where it is preprocessed and sent to the dataset collection module and respecting computing modules depending on the type of sensor. Wheel odometry, IMU, GNSS and tracking camera data is sent to the Perception module. Data from depth and RGB cameras is sent to the Jetson computing module for processing. Also, this data, along with the ultrasonic sensor readings, is transferred to the Point Cloud Maker module where the robot surroundings are modelled. Then all the processed data is sent to the Mapping Module that creates the complete map of the robot surroundings. This data is processed by the Local Path Planner module, and, along with the collision avoidance and Global Planner data, is sent to the PID Control module which, in turn, directly controls the wheels.

\section{Modification of Single Shot Detector}
As the default architecture for pedestrian detection, we chose the Single Shot MultiBox Detector (SSD) \cite{ssd} architecture with EfficientNet-B0 \cite{efnet} as the feature extraction backbone. These architectures were chosen because of their precision and computational efficiency.

\subsection{Single Shot Detector}
\begin{figure} [!h]
\centering
\includegraphics[width=0.6\linewidth]{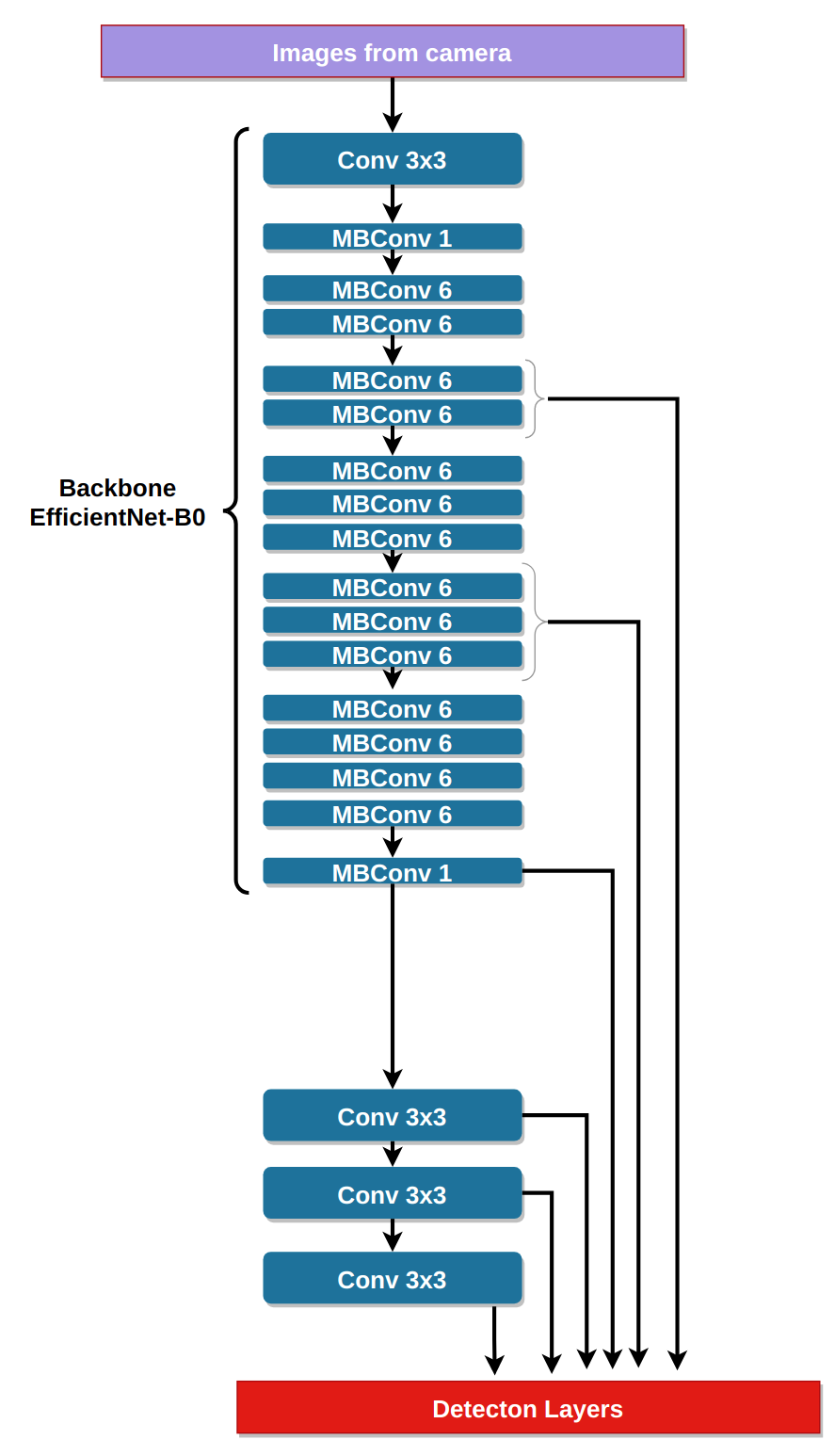}
\vspace{-1em}
\caption{Single Shot Detector architecture with EfficientNet-B0 feature extractor}
\vspace{-2em}
\label{fig4}
\end{figure}

The SSD architecture is based on a simple feed-forward neural network that creates a set of fixed-size bounding boxes and scores each box for the presence of class instances. Fig. \ref{fig4} shows the operating principles of SSD. First, the image is passed through a feature extractor (backbone), which extracts features at various convolutional layers. To acquire more spatial information about the image, additional convolutional layers extract more features for the detection block. This architecture does not use linear classification head because classification is performed along with the detection. 

All the features detected at various convolution layers are transferred to the object detection block. The detection is also carried out using convolution layers. Firstly, the features pass through 2 layers of convolution in parallel, one for classification for each prior box, and the other gives the probability of finding an object in each prior box. Then, numerous windows of various sizes and shapes is formed on the image. These windows are assumed to contain only one object, and a classifier is used to obtain the likelihood / estimate for each class. Once the detector generates a large number of bounding boxes, they must be limited to only one per object in the image. Non-Maximum Suppression (NMS), which is basically a form of a clustering algorithm, is the most commonly used approach for this task \cite{NMS}. 

\subsection{EfficientNet}
EfficientNet \cite{efnet} is one of state-of-the-art and fastest feature extraction backbones for classification models. The method of this neural network is based on the scaling of three parameters — the depth of the layers, the number of input and output channels and the spatial size. The previously developed methods scaled the dimension of the neural network arbitrarily (e.g., the number of layers and parameters). The proposed method in EfficientNet evenly scales parts of the neural network with fixed scaling factors. The efficiency of such scaling is directly influenced by the architecture of the network. To improve the performance of the neural network, the authors proposed to choose the initial architecture automatically using the neural architecture search framework. As a result, the initial model used the architecture with Inverted Residual blocks, similar to MobileNetV2 \cite{mobilenetv2} and MnasNet \cite{mnas}. The initial model was then scaled up and created the class of models called “EfficientNets”. 
 
The EfficientNet-B0, chosen as the backbone in this article, has a much smaller number of parameters, which allows it to run on mobile platforms.

\subsection{Dataset}
In our work, the SSD was trained using the PASCAL VOC dataset, which is considered one of the classic datasets for detection along with the COCO dataset for pedestrians detection and classification. From all classes in the dataset, only class “person” was used. In future work other classes, such as various animals, can also be added in the training since animals can appear on roads as well as humans. The architecture of this neural network was used as the baseline, and the performance of our modifications will be compared with this default model.

\subsection{Modified Single Shot Detector}
Despite being state-of-the art in computational efficiency, the chosen network architectures still struggle to provide real-time information on detected pedestrians, especially if the images have to be analyzed six times, once for each camera installed on the robot. To improve the speed of human detection algorithm, we propose an approach which eliminates the need to process a frame if no objects of interest are present in it. Since the situation when all the robot's cameras detect pedestrians at the same time is relatively infrequent, this approach will significantly improve the speed of the detection algorithm.

In order to achieve this task, we added another classification layer prior to the extra feature extraction convolutional layers. This classifier is trained for classification of two types of images: the ones containing humans and empty ones. After this classification, only images containing humans are sent to the SSD for detection (Algorithm \ref{alg1}). It is known that the computation complexity of 2D convolution layers depends on the number of $d$ tensor channels, kernel size $k$ and sequence length $n$ ($\mathcal{O}(k \cdot n \cdot d^2)$) \cite{complexity}; therefore, such modification should save computing power, since there will be no need to use additional convolutions for extra features and convolutions for detection in images not containing objects of interest.
\begin{algorithm}[!b]
\fontsize{10pt}{10.5pt}\selectfont
\SetAlgoLined
\KwData{Set of 6 image frames from cameras}
\KwResult{Bounding boxes and scores}
 \While{True}{
  frames preprocessing\;
  backbone layers forward\;
  \eIf{frame contains objects}{
   extra feature extraction layers forward\;
   detection layers forward\;
   frames postprocessing\;
   }{
   continue\;
  }
 }
 \caption{Modified SSD algorithm}
 \label{alg1}
\end{algorithm}

\begin{figure} [h]
\centering
\includegraphics[width=0.6\linewidth]{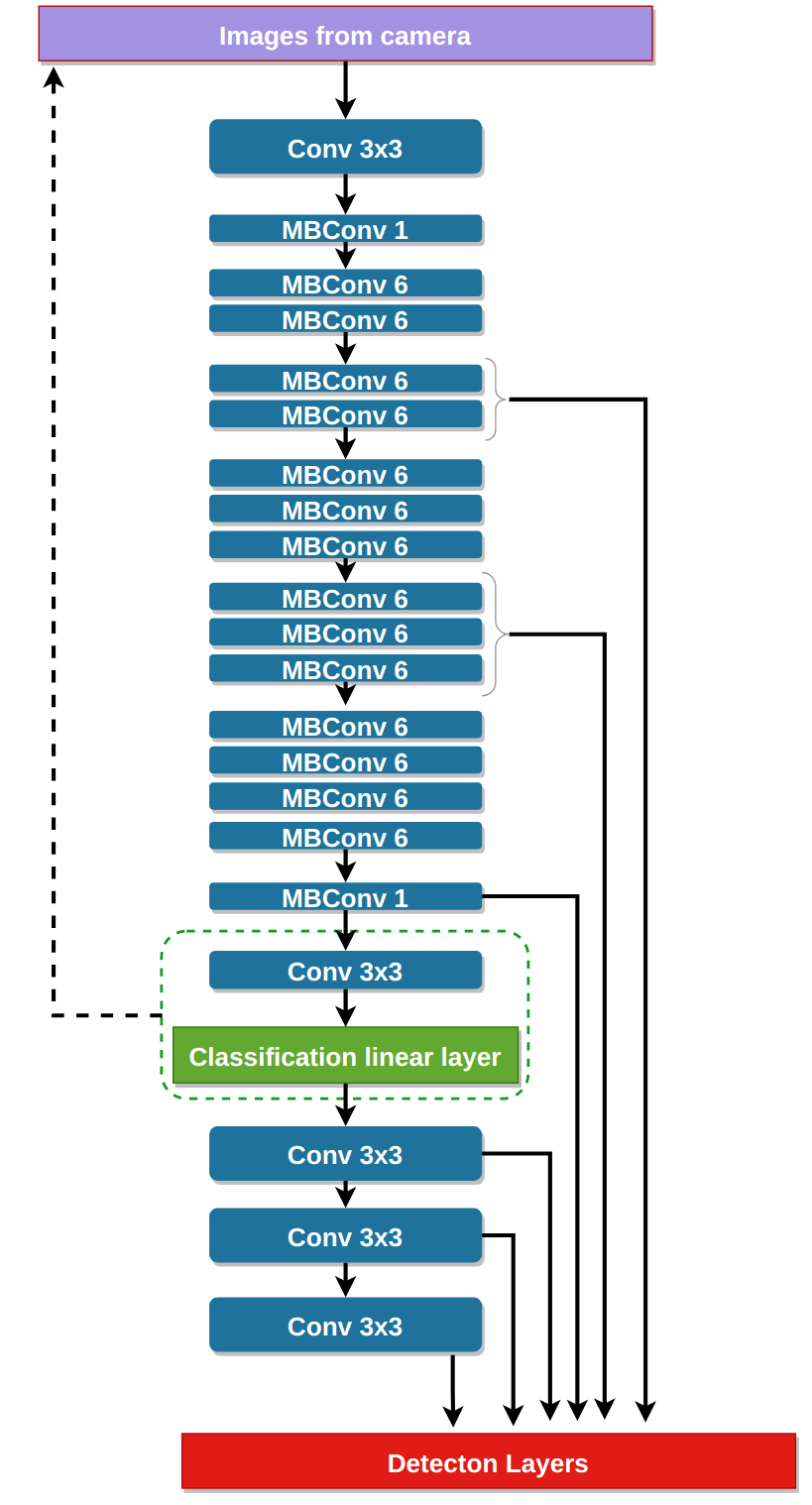}
\vspace{-1em}
 \caption{Modified Single Shot Detector's architecture}
 \vspace{-2em}
  \label{fig5}
\end{figure}

Fig. \ref{fig5} shows that the classification layer is located immediately after the convolution of the original backbone, which is structurally similar to the EfficientNet-B0 original classifier. The classification layer consists of a 1x1 convolutional layer, global pooling, and a dense layer with the output of 2 channels, which corresponds to the binary classification task. The original efficientNet-B0 has an input image size of 224x224, so, before entering the classification layer, the tensor had spatial dimensions of 7x7. SSD requires 300x300 images on input, so the same layer before classification will have other dimensions — 10x10.

\subsection{Loss function}
To train the SSD, the MultiBox loss was used, which is a linear combination of two losses as it was described in \cite{ssd}. For the loss function, the CrossEntropy loss was used, and, for convergence on bounding boxes, the usual smooth L1 Loss was used:
\begin{equation}
L_{\text {multibox}}=L_{\text {conf }}+\alpha \cdot L_{\text {loc }}.
\end{equation}
In our work, the image must be binary classified before entering the detection, so we need to use an additional loss in order for the neural network to be able to learn. We decided not to deviate from the pipeline and also use the linear combination of classification loss and MultiBox loss. The resulting loss looks as following:
\begin{equation}
L=L_{\text {multibox }}+\beta \cdot L_{\text {binary }}.
\end{equation}
Where $L_{\text {binary}}$ is binary cross entropy loss function for classification of human or non-human cases. In this case, $\beta$ is denoted as the hyperparameter for choosing the dimension of the loss function order.

\section{Experiments}
\subsection{Implementation Details}
The training of the SSD and the modified SSD happened for 300 epochs. The initial learning rate was equal to \textit{1e-3} and dropped \textit{10} times at epochs 200 and 270. The SGD optimizer was used with momentum \textit{0.9} and weight decay \textit{5e-4}.

\begin{table*}[!t]
\caption{Comparison of Modified SSD Performance in Different Cases}
\vspace{-1em}
\resizebox{2.05\columnwidth}{!}{%
\begin{tabular}{|c|c|c|cccc|cccc|c|}
\hline
\multirow{2}{*}{Case \#} & \multirow{2}{*}{\begin{tabular}[c]{@{}c@{}}Number of images \\ with object and \\ without it\end{tabular}} & \multirow{2}{*}{Equipment} & \multicolumn{4}{c|}{default SSD} & \multicolumn{4}{c|}{modified SSD} & \multirow{2}{*}{\begin{tabular}[c]{@{}c@{}}gain in \\ percent, \\ \%\end{tabular}} \\ \cline{4-11}
 &  &  & \multicolumn{1}{c|}{\begin{tabular}[c]{@{}c@{}}mean, \\ ms\end{tabular}} & \multicolumn{1}{c|}{\begin{tabular}[c]{@{}c@{}}min, \\ ms\end{tabular}} & \multicolumn{1}{c|}{\begin{tabular}[c]{@{}c@{}}max, \\ ms\end{tabular}} & \begin{tabular}[c]{@{}c@{}}SD, \\ ms\end{tabular} & \multicolumn{1}{c|}{\begin{tabular}[c]{@{}c@{}}mean, \\ ms\end{tabular}} & \multicolumn{1}{c|}{\begin{tabular}[c]{@{}c@{}}min, \\ ms\end{tabular}} & \multicolumn{1}{c|}{\begin{tabular}[c]{@{}c@{}}max, \\ ms\end{tabular}} & \begin{tabular}[c]{@{}c@{}}SD, \\ ms\end{tabular} &  \\ \hline
\multirow{2}{*}{1st case} & \multirow{2}{*}{\begin{tabular}[c]{@{}c@{}}0 image with object, \\ 6 without objects\end{tabular}} & PC with NVIDIA 1070 GPU & 49.80 & 49.58 & 50.66 & 0.25 & 38.5 &
38.19 & 39.65 & 0.31 & \textcolor{ForestGreen}{\textbf{-22.68\%}} \\ \cline{3-3}
 &  & Jetson Xavier & 167.76 & 164.48 & 193.30 & 4.45 & 127.11 & 125.48 & 135.67 & 2.65 & \textcolor{ForestGreen}{\textbf{-24.23\%}} \\ \hline
\multirow{2}{*}{2nd case} & \multirow{2}{*}{\begin{tabular}[c]{@{}c@{}}1 image with object, \\ 5 without objects\end{tabular}} & PC with NVIDIA 1070 GPU & 49.80 & 49.58 & 50.66 & 0.25 & 40.72 & 40.17 & 41.35 & 0.22 & \textcolor{ForestGreen}{\textbf{-18.24\%}} \\ \cline{3-3}
 &  & Jetson Xavier & 167.76 & 164.48 & 193.30 & 4.45 & 134.41 & 131.96 & 148.32 & 2.73 & \textcolor{ForestGreen}{\textbf{-19.88\%}} \\ \hline
\multirow{2}{*}{3rd case} & \multirow{2}{*}{\begin{tabular}[c]{@{}c@{}}2 image with object, \\ 4 without objects\end{tabular}} & PC with NVIDIA 1070 GPU & 49.80 & 49.58 & 50.66 & 0.25 & 42.61 & 42.34 & 43.62 & 0.30 & \textcolor{ForestGreen}{\textbf{-14.44\%}} \\ \cline{3-3} &  & Jetson Xavier & 167.76 & 164.48 & 193.30 & 4.45 & 142.83 & 140.30 & 158.59 & 3.31 & \textcolor{ForestGreen}{\textbf{-14.86\%}} \\ \hline
\multirow{2}{*}{4th case} & \multirow{2}{*}{\begin{tabular}[c]{@{}c@{}}3 image with object, \\ 3 without objects\end{tabular}} & PC with NVIDIA 1070 GPU & 49.80 & 49.58 & 50.66 & 0.25 & 44.40 & 44.14 & 45.15 & 0.24 & \textcolor{ForestGreen}{\textbf{-10.84\%}} \\ \cline{3-3}
 &  & Jetson Xavier & 167.76 & 164.48 & 193.30 & 4.45 & 150.30 & 146.49 & 168.36 & 4.37 & \textcolor{ForestGreen}{\textbf{-10.41\%}} \\ \hline
\multirow{2}{*}{5th case} & \multirow{2}{*}{\begin{tabular}[c]{@{}c@{}}4 image with object, \\ 2 without objects\end{tabular}} & PC with NVIDIA 1070 GPU & 49.80 & 49.58 & 50.66 & 0.25 & 46.24 & 45.89 & 46.92 & 0.23 & \textcolor{ForestGreen}{\textbf{-7.16\%}} \\ \cline{3-3}
 &  & Jetson Xavier & 167.76 & 164.48 & 193.30 & 4.45 & 156.94 & 153.91 & 181.71 & 4.35 & \textcolor{ForestGreen}{\textbf{-6.45\%}} \\ \hline
\multirow{2}{*}{6th case} & \multirow{2}{*}{\begin{tabular}[c]{@{}c@{}}5 image with object, \\ 1 without objects\end{tabular}} & PC with NVIDIA 1070 GPU & 49.80 & 49.58 & 50.66 & 0.25 & 48.21 & 47.87 & 49.11 & 0.32 & \textcolor{ForestGreen}{\textbf{-3.2\%}} \\ \cline{3-3}
 &  & Jetson Xavier & 167.76 & 164.48 & 193.30 & 4.45 & 163.07 & 158.98 & 201.62 & 6.41 & \textcolor{ForestGreen}{\textbf{-2.79\%}} \\ \hline
\multirow{2}{*}{7th case} & \multirow{2}{*}{\begin{tabular}[c]{@{}c@{}}6 image with object, \\ 0 without objects\end{tabular}} & PC with NVIDIA 1070 GPU & 49.80 & 49.58 & 50.66 & 0.25 & 50.82 & 50.56 & 51.91 & 0.26 & \textcolor{red}{+2.04\%} \\ \cline{3-3}
 &  & Jetson Xavier & 167.76 & 164.48 & 193.30 & 4.45 & 178.56 & 172.97 & 209.47 & 5.29 & \textcolor{red}{+6.44\%} \\ \hline
\end{tabular}
}
\vspace{-2em}
\end{table*}

\subsection{Inference experiments} 

\begin{figure} [h]
\vspace{-0.5em}
\centering
\includegraphics[width=1.0\linewidth]{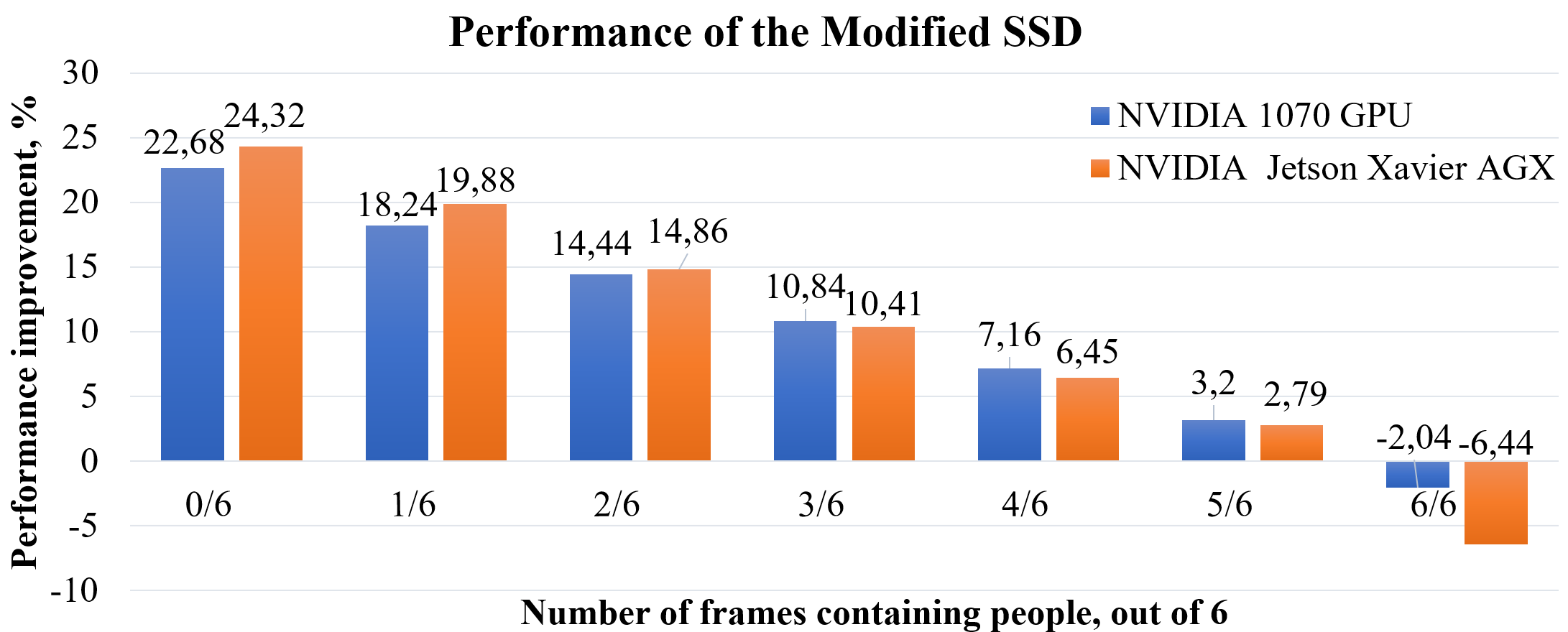}
\vspace{-1.5em}
\caption{Dependence of the change in the performance of the Modified SSD on the number of frames with people, compared to Classical SSD.}
\vspace{-1em}
\label{fig7}
\end{figure}

During the experiments, seven cases were examined with 0 to 6 camera readings containing target objects. For each of the cases, the experiment went as following: a sequence of 300 images (50 for each camera) containing or not containing target objects were sent to the input of the neural network, and the computation time was then averaged and compared with the computation time of the same images using the traditional network architecture. The experiments were conducted on the desktop computer with Nvidia 1070 GPU and Jetson Xavier computation module in order to compare desktop and mobile calculation times. The time taken to transfer images from RAM to GPU memory was not taken into account. In the experiment, only the GPU operating time was considered. The results of the experiments are presented in Table 1 and Fig. \ref{fig7}. Preliminary binary classification approach has shown significant decrease in the computation time for all cases except the one with all images containing target objects. However, the significant increase in processing time in other cases states that in certain environments the use of this method will significantly reduce the mean time for image processing.


\section{Conclusions}
We presented a novel approach to improving the computational efficiency of detection neural networks used in multi-camera setups. A lightweight classification network was implemented after the feature extraction step and before the detection step of the SSD neural network. A decision algorithm was implemented in order to skip images where no target objects were detected using the preliminary classifier. A set of experiments was conducted using different cases of target object presence for the computational efficiency validation of the proposed approach. The experimental results state that in most cases the proposed algorithm significantly reduces the computational complexity of object detection. The gain in computational efficiency reached more than 24\% with no objects present in the frames, and only decreased by 6\% when all cameras included target objects. This method is suitable for other detection architectures, where a classifier is used as a feature extractor. 

\section{Future Work}
Our future research will be devoted to determining the density of people around the robot in typical urban environments, such as city center, business districts, residential areas, and industrial areas. The varying pedestrian density in these areas may prove the efficiency of the proposed method in certain scenarios. It is also planned to test our method considering other important classes to detect (e.g., bicycles, E-scooters, dogs, cars).


\bibliographystyle{IEEEtran}
\bibliography{runbot}

\end{document}